# On reasoning in networks with qualitative uncertainty

Simon Parsons*and E. H. Mamdani
Department of Electronic Engineering,
Queen Mary and Westfield College,
Mile End Road,
London, E1 4NS, UK.

## Abstract

In this paper some initial work towards a new approach to qualitative reasoning under uncertainty is presented. This method is not only applicable to qualitative probabilistic reasoning, as is the case with other methods, but also allows the qualitative propagation within networks of values based upon possibility theory and Dempster-Shafer evidence theory. The method is applied to two simple networks from which a large class of directed graphs may be constructed. The results of this analysis are used to compare the qualitative behaviour of the three major quantitative uncertainty handling formalisms, and to demonstrate that the qualitative integration of the formalisms is possible under certain assumptions.

## 1 INTRODUCTION

In the past few years, the use of reasoning about qualitative changes in probability to deal with uncertainty has become widely accepted, being applied to domains such as planning [Wellman 1990b] and generating plausible explanations [Henrion and Druzdzel 1990]. Such a qualitative approach has certain advantages over quantitative methods, not least among which is the ability to model domains in which the relation between variables is uncertain as a result of incomplete knowledge, and domains in which numerical representations are inappropriate.

The existence of the latter, as Wellman [1990a] points out, is often due to the precision of numerical methods which can, in certain circumstances, lead to knowledge bases being applicable only in very narrow areas because of the interaction between values at a fine level of detail. Since they view the world at a higher level of abstraction, qualitative methods are immune to such problems; the small complications such interactions cause simply have no effect at the coarse level of detail with which qualitative methods are concerned.

The focus of the qualitative approach of Wellman and Henrion and Druzdzel is assessing the impact of evidence. That is assessing how the change in probability of one event due to some piece of evidence affects the probability of other events. For instance, taking a patient's temperature and finding that it is 38C is evidence that increases the probability that she has a fever, which in turn increases the probability that she has measles.

Now, when using the qualitative method we reason with a restricted set of values. Instead of using the full range of real numbers we are only interested in whether values are positive [+], negative [−], zero [0], or any of the three [?]. Thus we can determine that since the probability of fever increases, the change in probability is [+], and use this to decide that the change in probability of measles is also [+]. This is clearly weaker information than that obtained by traditional methods but may still be useful [Wellman 1990a], in particular since qualitative results may be obtained in situations where no numerical information may be deduced.

## 2 A NEW QUALITATIVE APPROACH

This paper presents a new approach to reasoning about qualitative changes. This work is drawn from the first author's thesis [Parsons 1993] in which may be found a number of extensions to the work described here. The motivation behind this work was to integrate different approaches to reasoning under uncertainty, in particular probability, possibility [Zadeh 1978] [Dubois and Prade 1988a], and evidence [Shafer 1976] [Smets 1988] theories. Thus, our qualitative approach differs from that described above in that it is concerned with changes in possibility values [Parsons 1992a] and belief values [Parsons 1992b] as well as probability values. As a result we need a general way of referring to values that may be probabilities, possibilities or beliefs.

---

*Current address: Advanced Computation Laboratory, Imperial Cancer Research Fund, P. O. Box 123, Lincoln's Inn Fields, London, WC2A 3PX, UK.

**Definition 2.1:** The certainty value of a variable $X$ taking value $x$, $val(x)$, is either the probability of $X$, $p(x)$, the possibility of $X$, $\Pi(x)$, or the belief in $X$, $bel(x)$.

We set our work in the framework of singly connected networks in which the nodes represent variables of interest, and the edges represent explicit dependencies between the variables. When the edges of such graphs are quantified with conditional probability values they are similar to those studied by Pearl [1988], when possibility values are used the graphs are similar to those of Fonck and Straszecka [1991] and when belief values are used the graphs are those studied by Smets [1991].

Each node in a graph represents a binary valued variable. The probability values associated with a variable $X$ which has possible values $x$ and $\neg x$ are $p(x)$ and $p(\neg x)$, and the possibility values associated with $X$ are $\Pi(x)$ and $\Pi(\neg x)$. Belief values may be assigned to any subset of the values of $X$, so it is possible to have up to three beliefs associated with $X$; $bel(\{x\})$, $bel(\{\neg x\})$ and $bel(\{x, \neg x\})$. For simplicity these will be written as $bel(x)$, $bel(\neg x)$ and $bel(x \cup \neg x)$. This rather restrictive framework is loosened in [Parsons 1993] where non-binary values and multiply connected are considered.

Wellman [1990a, 1990b] and Henrion and Druzdzel [1990] base their work upon the premise that a suitable interpretation of "$a$ positively influences $c$" is that:

$$p(c \mid a) \geq p(c \mid \neg a) \quad (1)$$

This seems reasonable, but it is a premise; there are other ways of encoding the information that seem equally intuitively acceptable, for instance $p(c \mid a) > p(c)$ and $p(c \mid a) > p(\neg c \mid a)$ [Dubois and Prade 1991]. Since our aim was to provide a method that was suitable for integrating formalisms we wanted to start from first principles thus minimising the number of necessary assumptions. As a result, a different approach was adopted as described below.

Given a link joining nodes $A$ and $C$ as in Figure 1, we are interested in the way in which a change in the value of $a$, say, expressed in a particular formalism, influences the value of $c$. Note that the arrow between $A$ and $C$ does not necessarily indicate a causal relationship between them, rather it suggests that propagation of qualitative changes will be from $A$ to $C$.

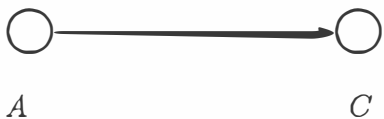

Figure 1: A simple network

We can then model the impact of evidence that affects the value of $A$ in terms of the change in certainty value of $a$ and $\neg a$, relative to their value before the evidence was known, and use knowledge about the way that a change in, say, $val(a)$ affects $val(c)$ to propagate the effect of the evidence. We define the following relationships that describe how the value of a variable $X$ changes when the value of a variable $Y$ is altered by new evidence:

**Definition 2.2:** The certainty value of a variable $X$ taking value $x$ is said to *follow* the certainty value of variable $Y$ taking value $y$, $val(x)$ follows $val(y)$, if $val(x)$ increases when $val(y)$ increases, and $val(x)$ decreases when $val(y)$ decreases.

**Definition 2.3:** The certainty value of a variable $X$ taking value $x$ is said to *vary inversely* with the certainty value of variable $Y$ taking value $y$, $val(x)$ varies inversely with $val(y)$, if $val(x)$ decreases when $val(y)$ increases, and $val(x)$ increases when $val(y)$ decreases.

**Definition 2.4:** The certainty value of a variable $X$ taking value $x$ is said to be *independent* of the certainty value of variable $Y$ taking value $y$, $val(x)$ is independent of $val(y)$, if $val(x)$ does not change as $val(y)$ increases and decreases.

The way in which the variation of $val(x)$ with $val(y)$ is determined is by establishing the qualitative value of the derivative $\partial val(x) \backslash \partial val(y)$ that relates them. If the derivative is known, it is a simple matter to calculate the change in $val(x)$ from the change in $val(y)$. Thus to determine the change at $C$ in Figure 1 we have:

$$\Delta val(c) = \Delta val(a) \otimes \left[\frac{\partial val(c)}{\partial val(a)}\right] \quad (2a)$$

$$\oplus \Delta val(\neg a) \otimes \left[\frac{\partial val(c)}{\partial val(\neg a)}\right]$$

$$\Delta val(\neg c) = \Delta val(a) \otimes \left[\frac{\partial val(\neg c)}{\partial val(a)}\right] \quad (2b)$$

$$\oplus \Delta val(\neg a) \otimes \left[\frac{\partial val(\neg c)}{\partial val(\neg a)}\right]$$

where $[x]$ is $[+]$ if $x$ is positive, $[\dot{0}]$ if $x$ is zero and $[-]$ if $x$ is negative, and $\oplus$ and $\otimes$ denote qualitative addition and multiplication respectively:

| $\oplus$ | [+] | [0] | [−] | [?] |
|---|---|---|---|---|
| [+] | [+] | [+] | [?] | [?] |
| [0] | [+] | [0] | [−] | [?] |
| [−] | [?] | [−] | [−] | [?] |
| [?] | [?] | [?] | [?] | [?] |

| $\otimes$ | [+] | [0] | [−] | [?] |
|---|---|---|---|---|
| [+] | [+] | [0] | [−] | [?] |
| [0] | [0] | [0] | [0] | [0] |
| [−] | [−] | [0] | [+] | [?] |
| [?] | [?] | [0] | [?] | [?] |

We can express this as a matrix calculation (after Far-

reny and Prade [1989]):

$$\begin{bmatrix} \Delta val(c) \\ \Delta val(\neg c) \end{bmatrix} = \begin{bmatrix} \left[\frac{\partial val(c)}{\partial val(a)}\right] & \left[\frac{\partial val(c)}{\partial val(\neg a)}\right] \\ \left[\frac{\partial val(\neg c)}{\partial val(a)}\right] & \left[\frac{\partial val(\neg c)}{\partial val(\neg a)}\right] \end{bmatrix} \quad (3)$$

$$\otimes \begin{bmatrix} \Delta val(a) \\ \Delta val(\neg a) \end{bmatrix}$$

Clearly $val(c)$ follows $val(a)$ when $\partial val(c)\backslash\partial val(a) = [+]$, $val(c)$ varies inversely with $val(a)$ when $\partial val(c)\backslash\partial val(a) = [-]$ and is independent of $val(a)$ when $\partial val(c)\backslash\partial val(a) = [0]$.

## 3 QUALITATIVE CHANGES IN SIMPLE NETWORKS

Applying probability theory to the example of Figure 1, so that $val(x)$ becomes $p(x)$ in (2) and (3), and referring to the directed link joining $A$ and $C$ as $A \rightarrow C$, we have the following simple result which agrees with the assumption (1) made by Wellman as a basis for his qualitative probabilistic networks when conditional values are taken as constant, as they are throughout this work:

**Theorem 3.1:** The relation between $p(x)$ and $p(y)$ for the link $A \rightarrow C$ is such that $p(x)$ follows $p(y)$ iff $p(x \mid y) > p(x \mid \neg y)$, $p(x)$ varies inversely with $p(y)$ iff $p(x \mid y) < p(x \mid \neg y)$ and $p(x)$ is independent of $p(y)$ iff $p(x \mid y) = p(x \mid \neg y)$ for all for all $x \in \{c, \neg c\}$, $y \in \{a, \neg a\}$.

**Proof:** Probability theory tells us that $p(c) = p(a)p(c \mid a) + p(\neg a)p(c \mid \neg a)$ and $p(a) = 1 - p(\neg a)$ so that $p(c) = p(a)[p(c \mid a) - p(c \mid \neg a)] + p(c\mid\neg a)$ and $[\partial val(c)\backslash\partial val(a)] = [p(c\mid a) - p(c\mid\neg a)]$. Similar reasoning about the way that $p(c)$ varies with $p(\neg a)$ and $p(\neg c)$ varies with $p(a)$ and $p(\neg a)$ gives the result. □

By convention [Pearl 1988] two nodes $A$ and $C$ are not connected in a probabilistic network if $p(a \mid c) = p(a \mid \neg c)$. In addition, since $p(a)$ and $p(\neg a)$ are related by $p(a) = 1 - p(\neg a)$, we can say that if $p(c)$ follows $p(a)$, then $p(\neg c)$ varies inversely with $p(a)$ and follows $p(\neg a)$. The assumption that conditional probabilities are constant does not seem to cause problems when propagating changes in singly connected networks as discussed here. However, the assumption does become problematic when handling multiply connected networks [Parsons 1993].

Applying possibility theory to the network of Figure 1, and writing $\Pi(x)$ for $val(x)$ in (2) and (3), we can establish a relationship between $\Pi(a)$ and $\Pi(c)$. Unfortunately, unlike the analogous expression for probability theory, this involves the non-conditional value $\Pi(a)$. This complicates the situation since the exact form of the qualitative relationship between $\Pi(a)$ and $\Pi(c)$ depends upon whether $\Pi(a)$ is increasing or decreasing. We have:

**Theorem 3.2:** The relation between $\Pi(x)$ and $\Pi(y)$, for all $x \in \{c, \neg c\}$, $y \in \{a, \neg a\}$, for the link $A \rightarrow C$ is such that $\Pi(x)$ follows $\Pi(y)$ if $min(\Pi(x \mid y), \Pi(y)) > min(\Pi(x \mid \neg y), \Pi(\neg y))$ and $\Pi(y) < \Pi(x \mid y)$. If $min(\Pi(x \mid y), \Pi(y)) \leq min(\Pi(x \mid \neg y), \Pi(\neg y))$ and $\Pi(y) < \Pi(x \mid y)$ then $\Pi(x)$ may follow $\Pi(y)$ up if $\Pi(y)$ is increasing, and if $min(\Pi(x \mid y), \Pi(y)) > min(\Pi(x \mid \neg y), \Pi(\neg y))$ and $\Pi(y) \geq \Pi(x \mid y)$ then $\Pi(x)$ may follow $\Pi(y)$ down if $\Pi(y)$ is decreasing. Otherwise $\Pi(x)$ is independent of $\Pi(y)$.

**Proof:** Possibility theory gives $\Pi(c) = sup[min(\Pi(c \mid a), \Pi(a)), min(\pi(c \mid \neg a), \Pi(\neg a))]$. This may not be differentiated, but consider how a small change in $\Pi(a)$ will affect $\Pi(c)$. If $\Pi(a)$ is the value that determines $\Pi(c)$, any change in $\Pi(a)$ will be reflected in $\Pi(c)$. This must happen when $min(\Pi(c \mid a), \Pi(a)) > min(\Pi(c \mid \neg a), \Pi(\neg a))$ and $\Pi(a) < \Pi(c \mid a)$. If $\Pi(a)$ is increasing and the second condition does not hold, it may become true at some point, and so the increase may be reflected in $\Pi(c)$. Similar reasoning may be applied when $\Pi(a)$ is decreasing and the first condition is initially false. Thus we can write down the conditions relating $\Pi(c)$ and $\Pi(a)$, while those relating $\Pi(c)$ and $\Pi(\neg a)$ as well as those relating $\Pi(\neg c)$ and $\Pi(a)$ and $\Pi(\neg a)$ may be obtained the same way. □

To formalise this we can say that $[\partial val(c)\backslash\partial val(a)] = [\uparrow]$ if $\Pi(x)$ may follow $\Pi(y)$ up and $[\partial val(c)\backslash\partial val(a)] = [\downarrow]$ if $\Pi(x)$ may follow $\Pi(y)$ down while extending $\otimes$ to give:

| $\otimes$ | [+] | [0] | [−] | [?] | [↑] | [↓] |
|---|---|---|---|---|---|---|
| [+] | [+] | [0] | [−] | [?] | [+, 0] | [0] |
| [0] | [0] | [0] | [0] | [0] | [0] | [0] |
| [−] | [−] | [0] | [+] | [?] | [0] | [−, 0] |
| [?] | [?] | [0] | [?] | [?] | [+, 0] | [−, 0] |

where $[+, 0]$ indicates a value that is either zero or positive. Normalisation, the possibilistic equivalent of $p(a) = 1 - p(\neg a)$, ensures that $max(\Pi(a), \Pi(\neg a)) = 1$. Thus at least one of $\Pi(a)$ and $\Pi(\neg a)$ is 1, and at most one of $\Pi(a)$ and $\Pi(\neg a)$ may change, so $\Pi(x)$ changes when either $\Pi(y)$ or $\Pi(\neg y)$ changes.

Writing $bel(x)$ for $val(x)$ in (2) and (3), and using Dempster's rule of combination [Shafer 1976] to combine beliefs in the network of Figure 1, we have:

**Theorem 3.3:** The relation between $bel(x)$ and $bel(y)$ for the link $A \rightarrow C$ is such that $bel(x)$ follows $bel(y)$ iff $bel(x \mid y) > bel(x \mid y \cup \neg y)$, $bel(x)$ varies inversely with $bel(y)$ iff $bel(x \mid y) < bel(x \mid y \cup \neg y)$ and $bel(x)$ is independent of $bel(y)$ iff $bel(x \mid y) = bel(x \mid y \cup \neg y)$ for all $x \in \{c, c\}$, $y \in \{a, a\}$.

**Proof :** By Dempster's rule $bel(c) = \sum_{a \subseteq \{a, \neg a\}} m(a)bel(c \mid a)$. Now, from Shafer [1976] $m(a) = bel(a)$, $m(\neg a) = bel(\neg a)$ and $m(a \cup \neg a) = 1 - bel(a) - bel(\neg a)$. Thus $\partial bel(c)\backslash\partial bel(a) = bel(c \mid a) - bel(c \mid a \cup \neg a)$. Sim-ilar reasoning about the way that $bel(c)$ varies with $bel(\neg a)$ and $bel(\neg c)$ varies with $bel(a)$ and $bel(\neg a)$ gives the result. □

Note that $bel(c)$ is the belief in hypothesis $c$ given all the available evidence, while $bel(c \mid a \cup \neg a)$ is the belief induced on $c$ by the marginalisation on $\{c \cup \neg c\}$ of the joint belief on the space $\{c \cup \neg c\} \times \{a \cup \neg a\}$. Thus $bel(c)$ follows $bel(a)$ if $c$ is more likely to occur given $a$ than given the whole frame. Other results are possible when alternative rules of combination, such as Smets' disjunctive rule [Smets 1991], are used.

The results presented in this section allow the propagation of changes in value from $A$ to $C$ given conditionals of the form $val(c \mid a)$. It is possible to derive similar results for propagation from $C$ to $A$ [Parsons 1993] which say, for instance, that if $p(c)$ follows $p(a)$, then $p(a)$ follows $p(c)$.

## 4 A COMPARISON OF THE THREE FORMALISMS

It is instructive to compare the qualitative behaviours of the simple link of Figure 1 when the conditional values that determine its behaviour are expressed using probability, possibility and evidence theories. This comparison exposes the differences in approach taken by the qualitative formalisms, providing some basis for choosing between them as methods of knowledge representation.

One way of representing the possible behaviours that a link may encode, is to specify the possible values of $\Delta val(\neg a)$, $\Delta val(c)$ and $\Delta val(\neg c)$ for given values of $\Delta val(a)$. Thus for probability theory we have:

$p(a) = 1$   If $\Delta p(a) = [0]$   Then $\Delta p(\neg a) = [0]$
            If $\Delta p(a) = [-]$   Then $\Delta p(\neg a) = [+]$
$p(a) \neq 1$ If $\Delta p(a) = [+]$   Then $\Delta p(\neg a) = [-]$
            If $\Delta p(a) = [0]$   Then $\Delta p(\neg a) = [0]$
            If $\Delta p(a) = [-]$   Then $\Delta p(\neg a) = [+]$

For any value of $p(a)$, either $[\partial p(c) \backslash \partial p(a)] = [+]$, or $[\partial p(c) \backslash \partial p(a)] = [-]$, and:

$p(c) = 1$   If $\Delta p(c) = [0]$   Then $\Delta p(\neg c) = [0]$
            If $\Delta p(c) = [-]$   Then $\Delta p(\neg c) = [+]$
$p(c) \neq 1$ If $\Delta p(c) = [+]$   Then $\Delta p(\neg c) = [-]$
            If $\Delta p(c) = [0]$   Then $\Delta p(\neg c) = [0]$
            If $\Delta p(c) = [-]$   Then $\Delta p(\neg c) = [+]$

The criterion on which the choice of probability theory is most likely to depend, is whether or not it is appropriate that $[\partial val(x) \backslash \partial val(\neg x)] = [-]$ in every case since it is possible to model this in other formalisms, and impossible to avoid it in probability theory.

In possibility theory we have:

$\Pi(a) = 1$   If $\Delta \Pi(a) = [0]$   Then $\Delta \Pi(\neg a) = [?]$
              If $\Delta \Pi(a) = [-]$   Then $\Delta \Pi(\neg a) = [+, 0]$
$\Pi(a) \neq 1$ If $\Delta \Pi(a) = [+]$   Then $\Delta \Pi(\neg a) = [0, -]$
              If $\Delta \Pi(a) = [0]$   Then $\Delta \Pi(\neg a) = [0]$
              If $\Delta \Pi(a) = [-]$   Then $\Delta \Pi(\neg a) = [0]$

For any $\Pi(a)$, either $[\partial \Pi(c) \backslash \partial \Pi(a)] = [0]$ or $[\partial \Pi(c) \backslash \partial \Pi(a)] = [+]$ while:

$\Pi(c) = 1$   If $\Delta \Pi(c) = [0]$   Then $\Delta \Pi(\neg c) = [?]$
              If $\Delta \Pi(c) = [-]$   Then $\Delta \Pi(\neg c) = [+, 0]$
$\Pi(c) \neq 1$ If $\Delta \Pi(c) = [+]$   Then $\Delta \Pi(\neg c) = [0, -]$
              If $\Delta \Pi(c) = [0]$   Then $\Delta \Pi(\neg c) = [0]$
              If $\Delta \Pi(c) = [-]$   Then $\Delta \Pi(\neg c) = [0]$

where $\Delta \Pi(x) = [?]$ is taken to mean $\Delta \Pi(x) = [+]$, $[0]$ or $[-]$. Thus possibility theory can represent a wider range of behaviours than probability theory.

However, possibility theory has one major limitation that is not shared by probability theory, and that is the fact that it does not have an inverting link. If $val(a)$ increases, it is only possible to have $val(c)$ decreasing if $val(\neg a)$ decreases and $val(c)$ follows it. This restricts the representation to the situation in which $val(a) \neq 1$ and increases to 1 and this may be inappropriate.

Evidence theory is the least restricted of the three. Here we have:

$bel(a) = 1$   If $\Delta bel(a) = [0]$   Then $\Delta bel(\neg a) = [?]$
              If $\Delta bel(a) = [-]$   Then $\Delta bel(\neg a) = [?]$
$bel(a) \neq 1$ If $\Delta bel(a) = [+]$   Then $\Delta bel(\neg a) = [?]$
              If $\Delta bel(a) = [0]$   Then $\Delta bel(\neg a) = [?]$
              If $\Delta bel(a) = [-]$   Then $\Delta bel(\neg a) = [?]$

For any $bel(a)$, $[\partial bel(c) \backslash \partial bel(a)] = [+]$, $[0]$, or $[-]$, while:

$bel(c) = 1$   If $\Delta bel(c) = [0]$   Then $\Delta bel(\neg c) = [?]$
              If $\Delta bel(c) = [-]$   Then $\Delta bel(\neg c) = [?]$
$bel(c) \neq 1$ If $\Delta bel(c) = [+]$   Then $\Delta bel(\neg c) = [?]$
              If $\Delta bel(c) = [0]$   Then $\Delta bel(\neg c) = [?]$
              If $\Delta bel(c) = [-]$   Then $\Delta bel(\neg c) = [?]$

so that there are no restrictions on the changes.

The purpose of this comparison is not to suggest that one formalism is the best in every situation. Instead, it is intended as some indication of which formalism is best for a particular situation. If a permissive formalism is required, then evidence theory may be the best choice, while probability might be better when a more restrictive formalism is needed.

## 5 QUALITATIVE CHANGES IN MORE COMPLEX NETWORKS

The analysis carried out in Section 3 allows us to predict how qualitative changes in certainty value will be propagated in a simple link between two nodes. Now, the change at $C$ depends only on the change at $A$, and differential calculus tells us that $\partial z \backslash \partial x = \partial z \backslash \partial y \cdot \partial y \backslash \partial x$ so the behaviours of such links may be composed. Thus we can predict how qualitative changes are propagated in any network, quantified by probabilities, possibilities or beliefs where every node has a single parent.

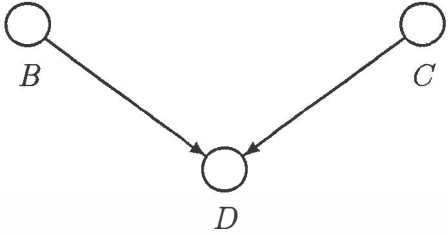

Figure 2: A more complex network

We now extend these results to enable us to cope with networks in which nodes may have more than one parent. To do this we consider the qualitative effect of two converging links such as those in Figure 2. Since we are only dealing with singly connected networks, B and C are independent and the overall effect at D is determined by:

$$\begin{bmatrix} \Delta val(d) \\ \Delta val(\neg d) \end{bmatrix} = \begin{bmatrix} \left[\frac{\partial val(d)}{\partial val(b)}\right] & \left[\frac{\partial val(d)}{\partial val(\neg b)}\right] \\ \left[\frac{\partial val(\neg d)}{\partial val(b)}\right] & \left[\frac{\partial val(\neg d)}{\partial val(\neg b)}\right] \end{bmatrix}$$
$$\otimes \begin{bmatrix} \Delta val(b) \\ \Delta val(\neg b) \end{bmatrix} \quad (4)$$
$$\oplus \begin{bmatrix} \left[\frac{\partial val(d)}{\partial val(c)}\right] & \left[\frac{\partial val(d)}{\partial val(\neg c)}\right] \\ \left[\frac{\partial val(\neg d)}{\partial val(c)}\right] & \left[\frac{\partial val(\neg d)}{\partial val(\neg c)}\right] \end{bmatrix}$$
$$\otimes \begin{bmatrix} \Delta val(c) \\ \Delta val(\neg c) \end{bmatrix}$$

There are two ways of tackling the network of Figure 2 in probability theory. We can either base our calculation on probabilities of the form $p(d \mid b)$ which implies the simplifying assumption that the effect of $B$ on $D$ is independent of the effect of $C$ (and vice versa), or we can use the proper joint probabilities of the three events $B$, $C$, and $D$, using values of the form $p(d \mid b, c)$.

In the first case we assume that the effects of $B$ and $C$ on $D$ are completely independent of one another so that the variation of $D$ with $B$ (and $D$ with $C$) is just as described by Theorem 3.1, the joint effect being established by using (4) to obtain:

$$[\Delta p(d)] = \left[\frac{\partial p(d)}{\partial p(b)}\right] \otimes [\Delta p(b)] \oplus \left[\frac{\partial p(d)}{\partial p(\neg b)}\right] \otimes [\Delta p(\neg b)]$$
$$\oplus \left[\frac{\partial p(d)}{\partial p(c)}\right] \otimes [\Delta p(c)] \oplus \left[\frac{\partial p(d)}{\partial p(\neg c)}\right] \otimes [\Delta p(\neg c)]$$

which gives the same results as the expression given by Wellman [1990a] for evaluating the same situation. With the other approach, writing the network as $B\&C \to D$, we have:

**Theorem 5.1:** The relation between $p(z)$ and $p(x)$ for the link $B\&C \to D$ is determined by:

$$\left[\frac{\partial p(z)}{\partial p(x)}\right] = [p(z \mid x, y) + p(z \mid \neg x, \neg y)]$$
$$- p(z \mid x, \neg y) - p(z \mid \neg x, y)]$$
$$\oplus [p(z \mid x, \neg y) - p(z \mid \neg x, \neg y)]$$

for all $x \in \{b, \neg b\}$, $y \{c, \neg c\}$ and $z \in \{d, \neg d\}$.

**Proof:** We have $p(d) = \sum_{b \in \{b, \neg b\} c \in \{c, \neg c\}} p(b, c, d) = \sum_{b \in \{b, \neg b\} c \in \{c, \neg c\}} p(b, c)p(d \mid b, c)$. Since $B$ and $C$ are independent, $p(b, c) = p(b)p(c)$. Using $p(x) = 1 - p(\neg x)$ and differentiating we find that $[\partial p(d) \backslash \partial p(b)] = p(c)[p(d \mid b, c) - p(d \mid \neg b, c)] + p(\neg c)[p(d \mid b, \neg c) - p(d \mid \neg b, \neg c)] = p(c)\{[p(d \mid b, c) + p(d \mid \neg b, \neg c)] - [(p(d \mid b, \neg c) + p(d \mid \neg b, c)]\} + [(p(d \mid b, \neg c) - p(d \mid \neg b, \neg c)]$. From this, and similar results for the variation of $p(d)$ with $p(\neg b)$, $p(c)$ and $p(\neg c)$, and the way $p(\neg d)$ changes with $p(b)$, $p(\neg b)$, $p(c)$ and $p(\neg c)$, the result follows. $\square$

Thus the way in which for instance $p(d)$ is dependent upon $p(b)$ is itself dependent upon a term just like the synergy condition introduced by Wellman, and applying (4) we get an expression which has a similar behaviour to that given by Wellman for a synergetic relation. In possibility theory we have a similar result to that for the simple link:

**Theorem 5.2:** The relation between $\Pi(x)$, $\Pi(y)$ and $\Pi(z)$, for all $x \in \{b, \neg b\}$, $y \in \{c, \neg c\}$, $z \in \{d, \neg d\}$ for the link $B\&C \to D$ is such that:

(1) $\Pi(z)$ follows $\Pi(x)$ iff $\Pi(x, y, z) > sup[\Pi(\neg x, y, z), \Pi(x, \neg y, z), \Pi(\neg x, \neg y, z)]$ and $\Pi(x) < min(\Pi(z \mid x, y), \Pi(y))$, or $\Pi(x, \neg y, z) > sup[\Pi(x, y, z), \Pi(\neg x, y, z), \Pi(\neg x, \neg y, z)]$ and $\Pi(x) < min(\Pi(z \mid x, \neg y), \Pi(\neg y))$.
(2) $\Pi(z)$ may follow $\Pi(x)$ up iff $\Pi(x, y, z) \leq sup[\Pi(\neg x, y, z), \Pi(x, \neg y, z), \Pi(\neg x, \neg y, z)]$ and $\Pi(x) < min(\Pi(z \mid x, y), \Pi(y))$, or $\Pi(x, \neg y, z) \leq sup[\Pi(x, y, z), \Pi(\neg x, y, z), \Pi(\neg x, \neg y, z)]$ and $\Pi(x) < min(\Pi(z \mid x, \neg y), \Pi(\neg y))$.
(3) $\Pi(z)$ may follow $\Pi(x)$ down iff $\Pi(x, y, z) > sup[\Pi(\neg x, y, z), \Pi(x, \neg y, z), \Pi(\neg x, \neg y, z)]$ and $\Pi(x) \geq min(\Pi(z \mid x, y), \Pi(y))$, or $\Pi(x, \neg y, z) > sup[\Pi(x, y, z), \Pi(\neg x, y, z), \Pi(\neg x, \neg y, z)]$ and $\Pi(x) \geq min(\Pi(z \mid x, \neg y), \Pi(\neg y))$.
(4) Otherwise $\Pi(z)$ is independent of $\Pi(x)$.

**Proof:** As for Theorem 3.2, the result may be determined directly from $\Pi(d) = sup_{x \in \{x, \neg x\}, y \in \{y, \neg y\}} \Pi(x, y, z)$ and $\Pi(x, y, z) = \Pi(z \mid x, y)\Pi(x)\Pi(y)$. $\square$

When we use belief values we may take the relationship between $B$, $C$ and $D$ to be determined by one set of conditional beliefs of the form $bel(d \mid b, c)$, or by two sets of conditional beliefs of the form $bel(d|b)$. For conditionals of the form $bel(d \mid b, c)$ we have:

**Theorem 5.3:** The relation between $bel(z)$ and $bel(x)$ for the link $B\&C \to D$ is determined by:

$$\left[\frac{\partial bel(z)}{\partial bel(x)}\right] = [bel(z \mid x, y) - bel(z \mid x \cup \neg x, y)]$$
$$\oplus [bel(z \mid x, \neg y) - bel(z \mid x \cup \neg x, \neg y)]$$

$$\oplus \, [bel(z \mid x, y \cup \neg y) - bel(z \mid x \cup \neg x, y \cup \neg y)]$$

For all $x \in \{b, \neg b\}$, $y \in \{c, \neg c\}$, $z \in \{d, \neg d\}$.

**Proof:** By Dempster's rule of combination, $bel(d) = \sum_{b \subseteq \{b, \neg b\}, c \subseteq \{c, \neg c\}} m(b)m(c)bel(d|b,c)$. Now, $m(x) = bel(x)$, $m(\neg x) = bel(\neg x)$ and $m(x \cup \neg x) = 1 - bel(x) - bel(\neg x)$, so that $[\partial bel(d) \backslash \partial bel(b)] = [(bel(d \mid b, c) - bel(d \mid b \cup \neg b, c)] \oplus [(bel(d \mid b, \neg c) - bel(d \mid b \cup \neg b, \neg c)] \oplus [(bel(d \mid b, c \cup \neg c) - bel(d \mid b \cup \neg b, c \cup \neg c)]$. From this, and similar results for the variation of $bel(d)$ with $bel(\neg b)$, $bel(c)$ and $bel(\neg c)$, and the way $bel(\neg d)$ changes with $bel(b)$, $bel(\neg b)$, $bel(c)$ and $bel(\neg c)$ the result follows. □

Thus $bel(d)$ follows $bel(b)$ iff $bel(d \mid b, c) > bel(d \mid b \cup \neg b, c)$, $bel(d \mid b, \neg c) > bel(d \mid b \cup \neg b, \neg c)$, and $bel(d \mid b, c \cup \neg c) > bel(d \mid b \cup \neg b, c \cup \neg c)$. For conditionals of the form $bel(d \mid b)$ we obtain:

**Theorem 5.4:** For the link $B \& C \rightarrow D$, $bel(z)$ follows $bel(x)$ if $bel(z \mid x) \geq bel(z \mid x \cup \neg x)$ and is indeterminate otherwise for all $x \in \{b, \neg b\}$, $y \in \{c, \neg c\}$, $z \in \{d, \neg d\}$.

**Proof:** Dempster's rule of combination tells us that $bel(d) = \sum_{b \subseteq \{b, \neg b\}, c \subseteq \{c, \neg c\}, b \lor c \supset d} bel(d|b)m(b) bel(d|c)m(c)$. As a result, $[\partial bel(d) \backslash \partial bel(b)] = [(bel(d \mid b) - bel(d \mid b \cup \neg b)] \{bel(d|\neg c)[1 + bel(c) + bel(\neg c) - m(c)] + bel(d|\neg c)[1 + bel(c) + bel(\neg c) - m(\neg c)] + bel(d|c)[1 + bel(c) + bel(\neg c) - m(c \cup \neg c)]\} + bel(d|\neg b)[m(c)bel(d \mid c) + m(\neg c)bel(d \mid \neg c) + m(c \cup \neg c)bel(d|c \cup \neg c))$. Since $m(x) \leq 1$ for all $x$, $[\partial bel(d) \backslash \partial bel(b)] = [+]$ if $bel(d \mid b) \geq bel(d|b \cup \neg b)$ and $[?]$ otherwise. From this, and similar results for the variation of $bel(d)$ with $bel(\neg b)$, $bel(c)$ and $bel(\neg c)$, and the way $bel(\neg d)$ changes with $bel(b)$, $bel(\neg b)$, $bel(c)$ and $bel(\neg c)$ the result follows. □

Thus the formalisms again exhibit differences in behaviour across the same network.

The expressions derived in this section are those obtained by using the precise theory of each formalism. This is important since it ensures the correctness of the integration introduced in Section 6. However, for reasoning using single formalisms, it may provde advantageous to extend the simpler apporach adopted by Wellman [1990a] to possibility and evidence theories.

Finally a word on the scope of the reasoning that we can perform as a result of our analysis. The differential calculus tells us that $\Delta z = \Delta x \cdot [\partial z \backslash \partial x] + \Delta y \cdot [\partial z \backslash \partial y]$, provided that $x$ is not a function of $y$. Thus we can clearly use the results derived above to propagate qualitative changes in probability, possibility and belief functions through any singly connected network.

## 6 INTEGRATION THROUGH QUALITATIVE CHANGE

The work described in this paper so far has extended qualitative reasoning about uncertainty handling formalisms to cover possibility and belief values as well as probability values. Not only is this useful in itself in providing a means of reasoning according to the precise rules of probability, possibility and Dempster-Shafer theory when there is incomplete numerical information, but it can also provide a way of integrating the different formalisms.

Consider the following medical example. The network of Figure 3 encodes the medical information that joint trauma ($T$) leads to loose knee bodies ($K$), and that these and arthritis ($A$) cause pain ($P$). The incidence of arthritis is influenced by dislocation ($D$) of the joint in question and by the patient suffering from Sjorgen's syndrome ($S$). Sjorgen's syndrome affects the incidence of vasculitis ($V$), and vasculitis leads to vasculitic lesions ($L$).

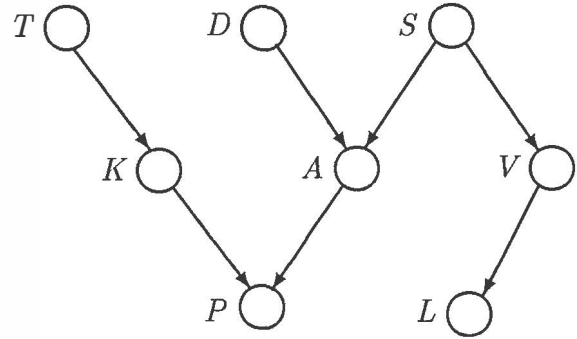

Figure 3: A network representing medical knowledge

The strengths of these influences are given as probabilities:

| | | | | | |
|---|---|---|---|---|---|
| $p(k \mid t)$ | = | 0.6 | $p(v \mid s)$ | = | 0.1 |
| $p(k \mid \neg t)$ | = | 0.2 | $p(v \mid \neg s)$ | = | 0.3 |
| $p(a \mid d, s)$ | = | 0.9 | $p(a \mid \neg d, s)$ | = | 0.6 |
| $p(a \mid d, \neg s)$ | = | 0.6 | $p(a \mid \neg d, \neg s)$ | = | 0.4 |

beliefs:

| | | |
|---|---|---|
| $bel(p \mid k, a)$ | = | 0.9 |
| $bel(p \mid k, \neg a)$ | = | 0.7 |
| $bel(p \mid \neg k, a)$ | = | 0.7 |
| $bel(p \mid k \cup \neg k, a)$ | = | 0.6 |
| $bel(p \mid k, a \cup \neg a)$ | = | 0.7 |
| $bel(\neg p \mid \neg k, \neg a)$ | = | 0.5 |
| $bel(\neg p \mid \neg k, a \cup \neg a)$ | = | 0.4 |

All other conditional beliefs are zero

and possibilities:

| | | | | | |
|---|---|---|---|---|---|
| $\Pi(l \mid v)$ | = | 1 | $\Pi(l \mid \neg v)$ | = | 1 |
| $\Pi(\neg l \mid v)$ | = | 0.1 | $\Pi(\neg l \mid \neg v)$ | = | 0.1 |

We can integrate this information allowing us to say how our belief in the patient in question being in pain,

and the possibility that the patient has vasculitic lesions, vary when we have new evidence that she is suffering from Sjorgen's syndrome. From the new evidence we have $\Delta p(s) = [+]$, $\Delta p(\neg s) = [-]$, $\Delta p(t) = [0]$, $\Delta p(\neg t) = [0]$, $\Delta p(d) = [0]$ and $\Delta p(\neg d) = [0]$. Since a change of $[0]$ can never become a change of $[+]$ or $[-]$ we can ignore the latter changes. Now, from Theorem 3.1 and Theorem 5.1 we know that:

$$\left[\frac{\partial p(v)}{\partial p(s)}\right] = [-] \qquad \left[\frac{\partial p(v)}{\partial p(\neg s)}\right] = [+]$$

$$\left[\frac{\partial p(a)}{\partial p(s)}\right] = [+] \qquad \left[\frac{\partial p(a)}{\partial p(\neg s)}\right] = [-]$$

so that $\Delta p(a) = [+]$, and $\Delta p(v) = [-]$ from which we can deduce that $\Delta p(\neg a) = [-]$ and $\Delta p(\neg v) = [+]$.

To continue our reasoning we need to establish the change in belief of $a$ and the change in possibility of $l$. To do this we make the monotonicity assumption [Parsons 1993] that if the probability of a hypothesis increases then both the possibility of that hypothesis and the belief in it do not decrease. As well as being intuitively acceptable, this assumption is the weakest sensible relation between values expressed in different formalisms, and is compatible both with the principle of consistency between probability and possibility values laid down by Zadeh [1978] and the natural extension of this principle to belief, necessity [Dubois and Prade 1988b], and plausibility values.

The assumption also says that if the probability decreases then the possibility and belief do not increase, and so we can say that $\Delta bel(a) = [+, 0]$, $\Delta bel(\neg a) = [-, 0]$, $\Delta \Pi(v) = [-, 0]$ and $\Delta \Pi(\neg v) = [+, 0]$. Now we apply Theorem 5.3 to find that:

$$\left[\frac{\partial bel(p)}{\partial bel(a)}\right] = [+] \qquad \left[\frac{\partial bel(p)}{\partial bel(\neg a)}\right] = [0]$$

$$\left[\frac{\partial bel(\neg p)}{\partial bel(a)}\right] = [-] \qquad \left[\frac{\partial bel(\neg p)}{\partial bel(\neg a)}\right] = [+]$$

Since we are initially ignorant about the possibility of vasculitis, we have $\Pi(v) = \Pi(\neg v) = 1$, so that Theorem 3.2 gives:

$$\left[\frac{\partial p(l)}{\partial p(v)}\right] = [0] \qquad \left[\frac{\partial p(l)}{\partial p(\neg v)}\right] = [0]$$

$$\left[\frac{\partial p(\neg l)}{\partial p(v)}\right] = [0] \qquad \left[\frac{\partial p(\neg l)}{\partial p(\neg v)}\right] = [0]$$

Hence we can tell that $\Delta bel(p) = [+, 0]$, $bel(\neg p) = [-, 0]$ and $\Delta \Pi(v) = \Delta \Pi(\neg v) = [0]$. The result of the new evidence is that belief in the patient's pain may increase, while the possibility of the patient having vasculitic lesions is unaffected. Thus we can use numerical values and qualitative relationships from different uncertainty handling formalisms to reason about the change in the belief of some event given information about the probability of a second event, and can infer whether the possibility of a third event also varies. As a result reasoning about qualitative change allows some integration between formalisms.

## 7 DISCUSSION

There is an important difference between the approach to qualitative reasoning under uncertainty described here, and that of Wellman [1990a, b]. Despite their name, Wellman's Qualitative Probabilistic Networks do not describe the qualitative behaviour of probabilistic networks exactly. In particular, some dependencies between variables are ignored in favour of simplicity, and synergy relations are sometimes introduced to represent them where it is considered to be important.

In our approach, since it is based directly upon the various formalisms, the qualitative changes predicted are exactly those of the quantitative methods. This has been demonstrated in [Parsons and Saffiotti 1993] which analyses the representation of a real problem in a number of different qualitative and quantitative formalisms. In this analysis we make qualitative predictions about the impact of evidence of faults in an electricity distribution network and compare these with the real quantitative changes. In every case, for probability, possibility and belief values, the qualitative predictions were correct. This verification is a good indication of the validity of the approach, and suggests that it will be useful in situations where incomplete information prevents the application of quantitative methods.

Our qualitative method also provides a means of integrating uncertainty handling formalisms on a purely syntactic basis. For any hypothesis $x$ about which we have uncertain information expressed, say in probability theory and possibility theory, we can make the intuitively reasonable assumption that if $p(x)$ increases $\Pi(x)$ does not decrease, and thus translate from probability to possibility without worrying what probability or possibility actually mean.

As a result any desired semantics may be attached to the values, a feature which finesses the problem of the acceptability of the semantics which must be faced by other, semantically based, schemes for integration (eg. [Baldwin 1991]). The only problem with switching semantics would be that some combination rules might no longer apply, Dempster's rule in the case of Baldwin's voting model semantics, which would entail a re-derivation of the appropriate propagation conditions. Since the qualitative approach does not a priori, rule out any combination scheme, this is not a major difficulty.

Finally, there is one important way that this method might be improved. The main disadvantage of any qualitative system is that there is no distinction between small values and large values, so 0.001 is qualitatively the same as 100, 000. As a result we cannot distinguish between evidence that induces small changes in the certainty of a hypothesis and evidence that induces large changes. This problem has been recognised for some time, and there is now a large body of work on order of magnitude reasoning (for example [Raiman 1986], [Parsons and Dohnal 1992]) which attempts to automate reasoning of the form it "If $A$ is bigger than $B$ and $B$ is bigger than $C$ then $A$ is bigger than $C$". The applications to our system are obvious, and we intend to do some work on this in the near future.

## 8 SUMMARY

This paper has introduced a new method for qualitative reasoning under uncertainty which is equally applicable to all uncertainty handling techniques. All that need be done to find the qualitative relation between two values is to write down the analytical expression relating them and take the derivative of this expression with respect to one of the values. This fact was illustrated by results from the qualitative analysis of the simplest possible reasoning networks in each of the three most widely used formalisms.

Having established the qualitative behaviours of probability, possibility and evidence theories, the differences between these behaviours were discussed at some length, before knowledge of this behaviour was used to establish a form of qualitative integration between formalisms. In this integration numerical and qualitative data expressed in all three formalisms was used to help derive the change in belief of one node in a directed graph and the possibility of another from knowledge of a change in the probability of a third, related, node.

### Acknowledgements

The work of the first author was partially supported by a grant from ESPRIT Basic Research Action 3085 DRUMS, and he is endebted to all of his colleagues on the project for their help and advice.

Special thanks are due to Mirko Dohnal, Didier Dubois, John Fox, Frank Klawonn, Paul Krause, Rudolf Kruse, Henri Prade, Alessandro Saffiotti and Philippe Smets for uncomplaining help and constructive criticism. The anonymous referees also made a number of useful comments.